**RESEARCH**

# Advances and Limitations in Open Source Arabic-Script OCR: A Case Study

Benjamin Kiessling[1], Gennady Kurin[2], Matthew Miller[2] and Kader Smail[2]

[1] Université Paris Sciences et Lettres, FR

[2] Faculty of Oriental Studies, University of Oxford, UK

Corresponding author: Matthew Miller (mtmiller@umd.edu)

This work presents an accuracy study of the open source OCR engine, Kraken, on the leading Arabic scholarly journal, al-Abhath. In contrast with other commercially available OCR engines, Kraken is shown to be capable of producing highly accurate Arabic-script OCR. The study also assesses the relative accuracy of typeface-specific and generalized models on the al-Abhath data and provides a microanalysis of the "error instances" and the contextual features that may have contributed to OCR misrecognition. Building on this analysis, the paper argues that Arabic-script OCR can be significantly improved through (1) a more systematic approach to training data production, and (2) the development of key technological components, especially multi-language models and improved line segmentation and layout analysis.

Cet article présente une étude d'exactitude du moteur ROC open source, Krakan, sur la revue académique arabe de premier rang, al-Abhath. Contrairement à d'autres moteurs ROC disponibles sur le marché, Kraken se révèle être capable de produire de la ROC extrêmement exacte de l'écriture arabe. L'étude évalue aussi l'exactitude relative des modèles spécifiquement configurés à des polices et celle des modèles généralisés sur les données d'al-Abhath et fournit une microanalyse des « occurrences d'erreurs », ainsi qu'une microanalyse des éléments contextuels qui pourraient avoir contribué à la méreconnaissance ROC. S'appuyant sur cette analyse, cet article fait valoir que la ROC de l'écriture arabe peut être considérablement améliorée grâce à (1) une approche plus systématique d'entraînement de la production de données et (2) grâce au développement de composants technologiques fondamentaux, notammentl'amélioration des modèles multilingues, de la segmentation de ligne et de l'analyse de la mise en page.

## Introduction

In late 2017 JSTOR initiated a collaboration with the Open Islamicate Texts Initiative (OpenITI) to run an Optical Character Recognition (OCR) pilot for the JSTOR Arabic digitization feasibility study (funded through a generous grant by the National Endowment for the Humanities).[1] The problem that JSTOR had encountered in their feasibility study was the problem that has long plagued efforts of scholars and librarians to digitize Arabic, Persian, Urdu, and Ottoman Turkish print documents: Arabic-script OCR programs produce notoriously poor results, despite the optimistic claims of some of their marketing materials (Alghamdi and Teahan 2017). Mansoor Alghamdi and William Teahan open their 2017 study by noting that "although handwritten script is significantly more challenging than printed Arabic text for OCR, Arabic printed text OCR still poses significant challenges." After evaluating Sakhr, Finereader, RDI Clever Page, and Tesseract (Version 3)—the main options for Arabic-script OCR—on Arabic print works they conclude that "all the evaluated Arabic OCR systems have low performance accuracy rates, below 75 per cent, which means that the time which would take to manually correct the OCR output would be prohibitive." These results are consonant with the authors' own experience using these OCR engines and those of our colleagues in the field of Islamicate Studies. In addition to these programs' lackluster performance, they also are not ideal systems for academic users for other reasons as well—for example, several are prohibitively expensive (for the average academic) and they offer little out of the box trainability (i.e., they come only with a generic OCR model and they cannot be trained to recognize new typefaces). With these problems in mind, in 2016 OpenITI began working on the development of open source OCR tools for Arabic-script languages (in print form) in collaboration with the computer scientist

1 OpenITI is a multi-institutional initiative that is focused on building digital infrastructure for the computational study of the texts of the Islamicate world. It is currently led by Dr. Matthew Thomas Miller (Roshan Institute for Persian Studies, University of Maryland, College Park), Dr. Sarah Bowen Savant (Aga Khan University, London), and Dr. Maxim Romanov (University of Vienna). Benjamin Kiessling (University of Leipzig/Université PSL) is one of OpenITI's primary computer science collaborators and he served as the technical lead for the OpenITI JSTOR OCR pilot. More information on JSTOR's NEH-funded project can be found in Kiplinger and Ray 2019.

Benjamin Kiessling. To date our work has primarily focused on Arabic-script OCR for print documents since handwritten text recognition (HTR) for Arabic-script manuscripts presents a series of additional issues (e.g., even more complex line segmentation and page layout analysis problems, a dizzying array of different script styles and scribal hands). However, we have begun preliminary experiments on Persian and Arabic manuscripts with some promising initial results using distantly supervised methods of training data production and the new line segmentation methods developed by Kiessling (Kiessling et al. 2019). (See also the experiments on HTR for Arabic manuscripts led by the British Library in Clausner et al. 2018; Keinan-Schoonbaert 2019; Keinan-Schoonbaert 2020). OpenITI's first OCR study with the new open source OCR engine, Kraken, developed by Kiessling, demonstrated that it was capable of achieving Arabic script-only accuracy rates >97.5% with as little as 800–1,000 lines of training data for that document's typeface (Kiessling et al. 2017). (Training data, in the context of OCR, consists of pairs of scans of individual lines of text with their digital transcription.) OpenITI has also replicated these high accuracy rates on Persian texts, with Perso-Arabic script-only accuracy rates ranging from 96.3% to 98.62% with typeface-specific models. (This work has not been published yet, but the full CER reports for these tests can be viewed in Open Islamic Texts Initiative 2021b, "OCR_GS_Data".) In this work, we present the results of our OCR study done in collaboration with JSTOR on the *al-Abhath* Arabic journal (arguably the most important Arabic language scholarly journal in the Middle East). In contrast with many OCR accuracy reports, in this study we performed both detailed manual and automatic Character Error Rate (CER) accuracy checks, which enabled us to develop a much more fine-grained understanding of where the Kraken OCR engine was failing to properly transcribe the Arabic text. These results confirm Kraken's ability to produce highly accurate Arabic-script OCR, but they also provide new insights into the importance of systematic training data production, the relative accuracy of typeface-specific and generalized models, and the key technological improvements needed for improved Arabic-script OCR output.

Section two reviews the open-source software used in this study before the JSTOR Arabic OCR pilot and accuracy study are described in sections three and four, respectively. Section five contains our recommendations for the necessary technological and data improvements needed to improve Arabic OCR in the future.

## OpenITI OCR software: Kraken

Kraken is an open-source OCR engine that was developed by Benjamin Kiessling. It utilizes a segmentationless sequence-to-sequence approach (Graves et al. 2006) with a hybrid convolutional and recurrent neural network to perform character recognition (Kiessling 2019) which obviates the need for character-level segmentation—i.e., the neural network responsible for text extraction from image data recognizes whole lines of text without resorting to smaller subunits like words or characters which can be difficult to accurately compute for languages written in connected script.

As an initial preprocessing step page images are converted into black and white through a process called binarization. Layout analysis, i.e. the detection of lines for subsequent steps, is then performed on this binarized image with an algorithm based on traditional computer vision methods. In a final step, the previously detected rectangular lines are fed into the neural network for character recognition. (More on Kraken's technical details can be found in Kiessling, n.d.) The benefit of eliminating fine-grained segmentation in comparison to older character segmenting systems such as tesseract 3, Sakhr, and most likely Abbyy FineReader (as a proprietary software the exact nature of the classifier is unknown) can be seen not only during recognition but also in the streamlined production process of training data for adaptation of the OCR system to new scripts and typefaces. With character-based systems annotators have to manually locate and transcribe single characters while Kraken is trained on full line transcriptions which are faster to annotate and verify, especially in the case of connected scripts.

## The OpenITI JSTOR OCR pilot

OpenITI began the JSTOR OCR pilot by performing a randomized review of the Arabic typefaces used in each year of the *al-Abhath* journal. The page images of the journal were obtained from the Arabic and Middle Eastern Electronic Library (Project AMEEL) of Yale University Library. It was determined that there were two basic typefaces in the *al-Abhath* journal archive, with the first typeface being much more prevalent than the second (typeface #1: volumes 1–33, 36–39, 48–50; typeface #2: volumes 34–35, 40–47). Examples of these two typefaces can be seen in **Figures 1–2** (for comparison sake, the last word in both lines—on the left of the page—is the same word).

Both typefaces had some internal font differences and other minor character/script variations (e.g., patterns of use of *alef hamza*, slight shifts in placement of dots, slight differences in the degree of curvature of lines in a couple of instances, and minor ligature differences). This intra-typeface variation was especially apparent in typeface #1, which had a long run as *al-Abhath's* typeface. To address this issue it was decided that the best approach would be to produce approximately 5,000 lines of training data for the first typeface and 2,000 lines of training data for the second typeface.

After a randomized sample of the pages representing each typeface were selected, research assistants working with the OpenITI team produced the training data for these 7,000 lines using CorpusBuilder 1.0—a new OCR postcorrection platform produced through the collaboration of OpenITI and Harvard Law School's SHARIAsource project. (For more on CorpusBuilder 1.0, please see Open Islamicate Texts Initiative, "CorpusBuilder 1.0," 2021a) After these 7,000 lines of training data were double checked for accuracy, a final spot review was conducted. This "gold standard" data was then utilized for model production and OCR. (This training data is available for reuse and can be found in Open Islamic Texts Initiative (2021b, "OCR_GS_Data")

The first round of OCR accuracy tests was performed by an outside contractor, which JSTOR hired to conduct a ten-page manual accuracy comparison between the Kraken output and the corresponding output for ABBYY (see **Table 1**).

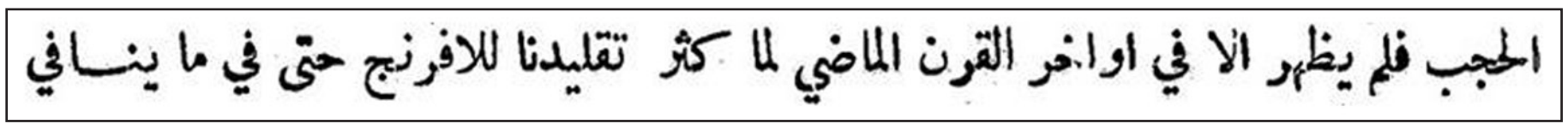
الحجب فلم يظهر الا في اوا.خر القرن الماضي لما كثر تقليدنا للافرنج حتى في ما ينـــافي

**Figure 1:** Typeface #1 Sample.

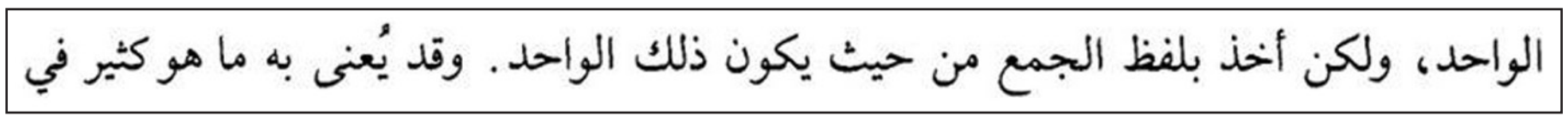
الواحد، ولكن أخذ بلفظ الجمع من حيث يكون ذلك الواحد. وقد يُعنى به ما هو كثير في

**Figure 2:** Typeface #2 Sample.

**Table 1:** Contractor's Accuracy Comparison of Abbyy and OpenITI (Kraken) OCR Results.

| Page (Tiff) Number | Total Number of characters | Abbyy Character Errors | OpenITI Character Errors | Abbyy Accuracy Rate | OpenITI Accuracy Rate |
|---|---|---|---|---|---|
| **Page #1 (00010004_ 187997831.tif)** | 1230 | 270 | 38 | 78.049% | 96.911% |
| **Page #2 (00010004_ 187997832.tif)** | 37 | 15 | 27 | 59.459% | 27.027% |
| **Page #3 (00010031_ 187998459.tif)** | 3182 | 355 | 23 | 88.843% | 99.277% |
| **Page #4 (00010063_ 187999338.tif)** | 3157 | 327 | 29 | 89.642% | 99.081% |
| **Page #5 (00010129_ 188001031.tif)** | 3222 | 378 | 16 | 88.268% | 99.503% |
| **Page #6 (00010012.tif)** | 3259 | 326 | 75 | 89.997% | 97.699% |
| **Page #7 (00010030.tif)** | 2503 | 230 | 17 | 90.811% | 99.321% |
| **Page #8 (00010126.tif)** | 2631 | 252 | 170 | 90.422% | 93.539% |
| **Page #9 (00010127.tif)** | 2294 | 223 | 35 | 90.279% | 98.474% |
| **Page #10 (00010132.tif)** | 2296 | 243 | 96 | 89.416% | 95.819% |

With the exception of page #2, OpenITI performed substantially better on the pages the contractor reviewed, achieving >99% accuracy in 4/10 pages, >97% accuracy in 6/10 pages, >95.8% in 8/10 pages. The exceptions to these generally impressive numbers were pages #2 and #8 in which OpenITI only achieved 27.027% and 93.539% respectively. While the contractor's review was quite useful and generally confirmed OpenITI's results from its previous work (i.e., that Kraken achieves significantly higher accuracy rates on Arabic texts than the commercial OCR solutions for Arabic), the OpenITI team discovered upon further review that there were several problems with the contractor's study. (For OpenITI's previous study on *Kraken*, please see: Kiessling et al. 2017.)

First, Page #2—by far the most disappointing result—is a highly atypical page of *al-Abhath* data. It only contains 37 characters total and much of these are contained in a large header that is in a highly calligraphic script and is heavily vocalized (see **Figure 3**). It is noteworthy that Abbyy performed better on this script, but this page is an extreme outlier in the data.

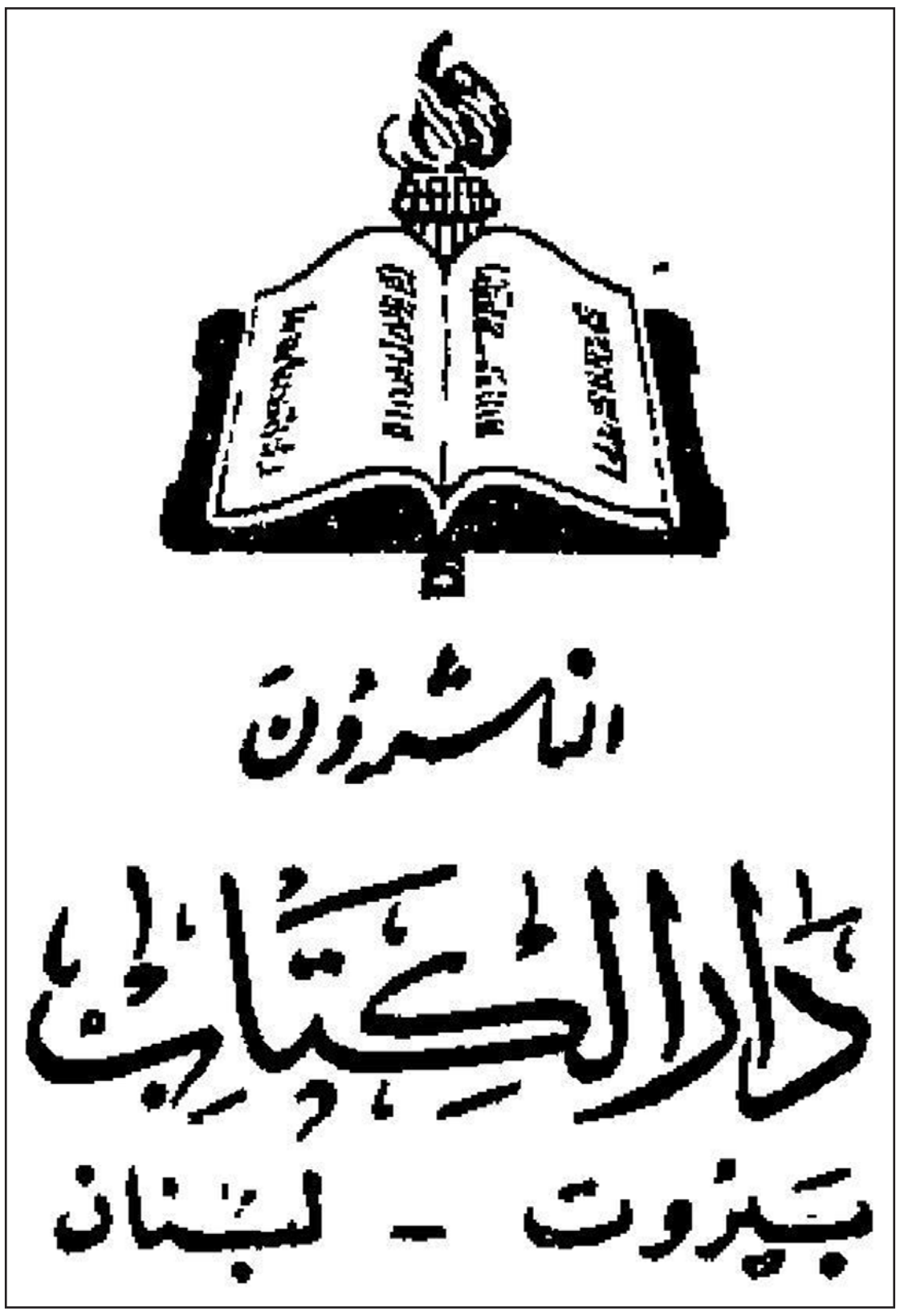

**Figure 3:** Header.

The second issue that we identified in the contractor's review was that they were marking certain differences between the original scans and OCR output as errors which were not true errors, and, in some cases, even marked some characters in the OCR output as errors that were not errors at all. For example, in the former case, they marked all numbers as errors in the OpenITI OCR output which were rendered as western Arabic numerals (e.g., 1, 2, 3) instead of as eastern Arabic numerals (e.g., ١, ٢, ٣)—a problem that was particularly prevalent on page #8 (thus at least partially responsible for OpenITI's comparatively lower accuracy rate on page #8). The OCR rendered them as western Arabic numerals instead of eastern Arabic numerals because we decided to merge western and eastern Arabic numerals into their universal numerical values in the OCR process and then represent that value in western Arabic numerals in the OCR output. (This practice of collapsing numeric values to their universal numerical value can be done for multiple reasons, but, in this particular case, one of our primary motivating factors was the fact that there were inconsistencies in the transcription practice of numbers in the training data.) These differences, thus, are not true errors—their numerical value is correct—and their representation can be changed to eastern Arabic numerals if that is what users prefer. Another similar issue was discovered in the contractor's treatment of diacritics: they routinely marked correctly rendered words as incorrect if the word's original diacritics were not included in the OCR output. However, again, this difference in the original text and the OCR output is not a true error in transcription because OpenITI has followed the practice (with one exception discussed further below) of not reproducing vocalization in its training data (for reasons elaborated below) and thus the fact that vocalization were not rendered in OpenITI's OCR output is actually a sign that the Kraken OCR engine was functioning correctly. (This training data generation practice can be changed if the users desire, and given the results in OpenITI's larger accuracy study described below, this change may be advisable in the future, depending on the requirements of each individual user's use case.)

These problems in the contractor's approach to error designation led them to calculate lower accuracy estimates for OpenITI OCR output than it achieved in actuality—a problem that was particularly accentuated in the case of page #8, which contained a larger amount of numbers than the other pages the contractor reviewed. Due to the problems discovered in the contractor's initial accuracy study, JSTOR requested that OpenITI perform a more detailed accuracy assessment on approximately fifty pages.

## OpenITI accuracy study

The OpenITI team began by generating automatic character error rate (CER) reports for the *al-Abhath* data (see **Table 2** for full results). In the first round of experiments, we built two different models—typeface model #1 and #2—based on the two different sets of training data produced for the two typefaces that we identified in the full run of *al-Abhath*. After extracting 1,000 lines of training data from the original 5,000 lines for typeface #1 and 700 lines of training data from the original 2,000 lines for

**Table 2:** Overview of OCR accuracy rates (drawn from character error rate (CER) reports).

| Model | Total Character Accuracy | Arabic Script Only Accuracy | Common Character Accuracy* | Inherited Character Accuracy** |
|---|---|---|---|---|
| **Typeface #1 Model** | 95.96% | 97.56% | 96.91% | 79.67% |
| **Typeface #2 Model** | 94.84% | 97.11% | 94.16% | 85.18% |
| **Generalized Model** | 97.41% | 98.46% | 96.36% | 89.44% |

* "Common characters" are characters shared by multiple scripts, primarily punctuation and other signs and symbols. In Arabic script, the *kashīda* or *tatwīl* (elongation character) is included in the common script class.

** "Inherited characters" are characters, such as vocalization, that can be used on multiple languages and they only come to be defined in reference to the character with which they are combined (i.e., they "inherit" the script of the base character with which it is used).

typeface #2 to use as validation sets, we then trained the model on the remaining lines and tested these models' accuracy using the validation sets. This method of isolating a fixed number of lines of the training data as a validation set for automatic accuracy testing is a standard procedure when evaluating machine learning models. These accuracy results can be found in rows 2–3 in **Table 2**. These typeface-specific models were the ones used to produce the OCR output that was transferred to JSTOR and that the contractor reviewed for their accuracy study.

In the time between the delivery of the *al-Abhath* OCR output to JSTOR and OpenITI's manual accuracy study (discussed below), we began developing a generalized Arabic model from all of the training data that OpenITI has produced over the last year two years—circa 15,000 lines. (All of this gold standard training data can be found in Open Islamic Texts Initiative 2021b, "OCR_GS_Data.") Generalized OCR models incorporate character features from all of the typefaces represented in the data upon which it is trained and therefore can often achieve higher levels of accuracy on a broader range of typefaces. We decided to test this model on all of the *al-Abhath* data to determine if total OCR accuracy could be improved and, if so, by how much. The results, shown in row #4 of **Table 2**, were impressive. For this accuracy assessment, 2,096 lines of the 7,000 lines of training data were isolated as a validation set. The generalized model's total character accuracy rate was 97.41%—a 2.57% improvement over the typeface #2-only model (i.e., a ~50% improvement rate) and 1.45% improvement over the typeface #1-only model—and its Arabic script-only accuracy went up to a respectable 98.46%. The generalized model performed better than the typeface-specific models in all categories, but its most significant gains were in the category of "inherited" characters.

According to the CER reports, the most significant source of errors in both the typeface #2 model and the generalized model were whitespace (spacing) errors and the Arabic vocalization marker, *fatḥa tanwīn* (unicode codepoint: Arabic fatḥatan). In the case of the typeface #1 model, whitespace errors were again the most significant source of errors, followed by *kāf* (ك), *yā'* (ي), and then *fatḥa tanwīn* errors. The *hamza above* (ٔ) character ranks as the seventh most common error in typeface #1 model and fifth in typeface #2 model. The *mīm* (م) character also is a common error in both the typeface #1 and #2 models, ranking as the sixth and fifth most common error in their CER reports respectively.

Concurrent with the generation of CER reports, the OpenITI team began a far more expansive manual review of fifty—randomly selected—pages of the original OCR output produced by the typeface #1 and typeface #2-specific models. Each of these fifty pages were reviewed and then their error reports were collated into a master list of 1,096 total error instances. We use the term "error instance" here to highlight the fact that we are not exclusively recording individual, one-to-one character errors, but instances in the text in which one or more characters were read incorrectly. In most cases, this is a one-to-one character mistranscription, but in some other cases one character in the original was read as two or more in the OCR output or multiple characters in the original were read as one or none in the OCR output. In a few cases—discussed in more detail below—there are whole sections of text that are severely mistranscribed due to one or another feature in the original text. Finally, each error instance was examined with an eye towards identifying possible factors in its adjacent context that may have led to that error and then coded with any of the following categories that were applicable:

1. ***Poor scan quality:*** an element in the raw scan is unclear, or extraneous marks are present.
2. ***Ligature/atypical letter or dot form:*** connection between letters or placement of dots is in a less common form.
3. ***Vocalization:*** vocalization marks were present in original word.
4. **Kashīda/tatwīl** ***(elongation character):*** error appears in the context of a word that has been elongated.
5. ***Header/font alteration:*** bolded, italicized, or enlarged text.
6. ***Footnote:*** error appears in the context of a footnote.
7. ***Format:*** atypical format of presentation, e.g., table, list.
8. ***Hamza:*** mistranscribed character was a *hamza* or a *hamza* was present in the original word that was mistranscribed.
9. ***Doubled character:*** a single letter or number in the original scan was doubled in the OCR output.
10. ***Missed*** **fatḥa tanwīn*****:*** *fatḥa tanwīn* in the original text was not transcribed.
11. ***Punctuation or other symbol:*** error was a punctuation mark or other symbol.
12. ***Non-Arabic language:*** original text was not Arabic.
13. ***Numbers:*** error was a number.
14. ***Superscript numerals:*** error was a footnote numeral in the body of the text.

This list of error codes is a mixture of error types (#8–14) and the most common recurring contextual features of the errors (#1–7). For categories of the latter type, it is important to emphasize that the presence of any of these contextual features near an error in the original text does not necessarily mean that it *caused* the error. But their repeated co-occurrence may be related and thus suggest future avenues of research and/or the need to better address this issue in the process of future training data production. We should also point out that in the case of some errors none of the following category codes were applicable, which only means that the reason for their improper rendering was not immediately evident to the human reviewers.

**Table 3**: Error coding for error instances in OpenITI manual OCR output assessment.

| Error Code | Quantity Identified |
|---|---|
| Poor scan quality | 25 |
| Ligature/atypical letter or dot form | 182 |
| Vocalization | 90 |
| *Kashīda/tatwīl* (elongation character) | 31 |
| Header/font alteration | 113 |
| Footnote | 88 |
| Format | 14 |
| *Hamza* | 97 |
| Doubled letter | 209 |
| Missed *fatḥa tanwīn* | 91 |
| Punctuation or other non-alphanumeric symbols | 25 |
| Non-Arabic language | 70 |
| Numbers | 94 |
| Superscript numerals | 26 |

We do want to preface our presentation of the results of this manual review and error coding below with one further cautionary note. Manual evaluations are both essential and problematic: they provide far more detailed data (i.e., "thick data") about the OCR output and where OCR is failing, but they are much more time and labor intensive (and thus more limited in scope) and subject to human error. The results presented in **Table 3** should be understood in this light. They should be understood as a snapshot of the human-inferable errors present in the OCR output. Each error type in **Table 3** and possible ways to address it will be discussed in more detail in separate sections.

## Doubled letter

The "doubled letter" error type was the most frequent that we observed in the OCR output data (see example in **Figure 4**). (In the images in **Figures 4–31** the Arabic text at the top of the images is the original scan and the text below is the OCR output.)

At first this error was perplexing. However, it was subsequently discovered that these "doubling" errors were an artifact of the decoding algorithm converting the sequence of confidences for each character produced by the neural network into a series of characters. As the network assigns each character a probability for each pixel-wide vertical slice of the input line image and printed characters are wider than a single pixel an algorithm is needed to extract the actual line text from the longer character probability sequence. Our implementation was based on a thresholding and merging approach which can cause doubled characters when the network assigns a probability below the threshold for a character at a vertical slice between high probability slices for the same character. As a simplified example, assume the network assigns a probability for character x at 4 vertical slices: (0.9, 0.95, 0.6, 0.9). Decoding with a threshold of 0.5 will result in an intermediate sequence xxxx that is then merged to x, while selecting a higher threshold of 0.7 will result in a potentially erroneous character sequence of xx merged from xx_x. This error has been effectively addressed by switching to a greedy decoding which always uses the highest probability character at each vertical slice.

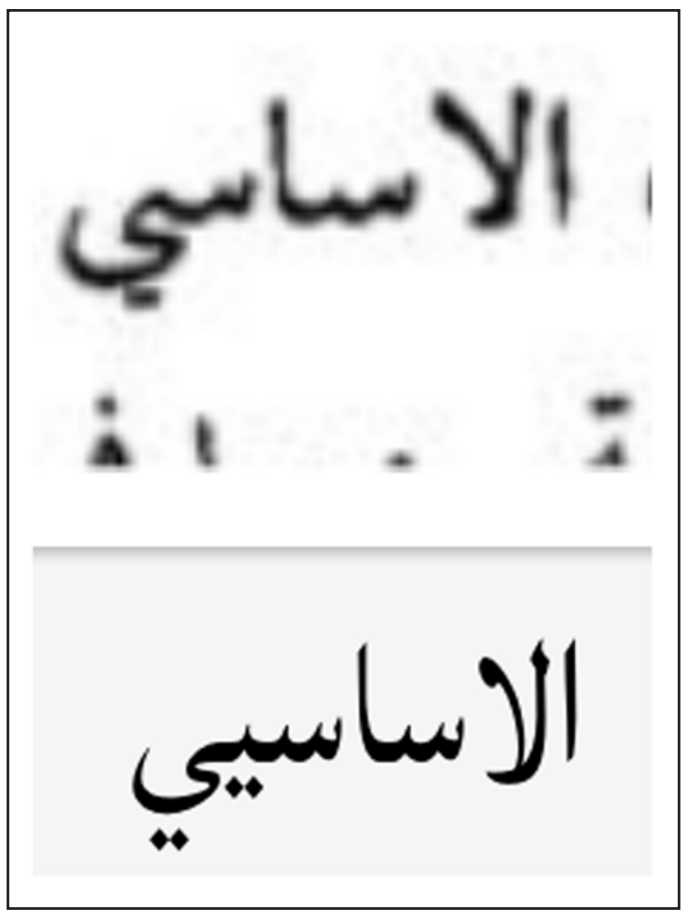

**Figure 4:** "Doubled Letter" errors.

## Header/Font alteration, footnotes, and superscript numerals

Errors that occurred in the context of changes in the font (bolded, italicized, enlarged/decreased text size) represent the largest category of errors in the OCR output. Their total numbers are not even fully reflected in **Table 3** because examples in which whole sections (see examples in **Figures 5–6**) were severely mistranscribed were not enumerated (character-by-character) in the error totals of the OpenITI manual assessment. Such sections were very rare, and in most other cases the OCR still rendered text with font alterations with a relatively high degree of accuracy, but font alterations do seem to increase error rates. One of the most common examples of this issue was observed in text headers (including both section headers and chapter titles), which were typically bolded or bolded and enlarged in *al-Abhath* (see **Figures 7–8**). In some headers, as mentioned above, an entirely different typeface was employed (see **Figure 3**)—although this is a less common practice.

Other modifications to the font of the typeface, e.g., footnotes, superscript (decreased text size) numerals, also seem to be correlated with decreased accuracy rates. In the future, this could be addressed by ensuring that a sufficient number of lines of training data with such font modifications is included.

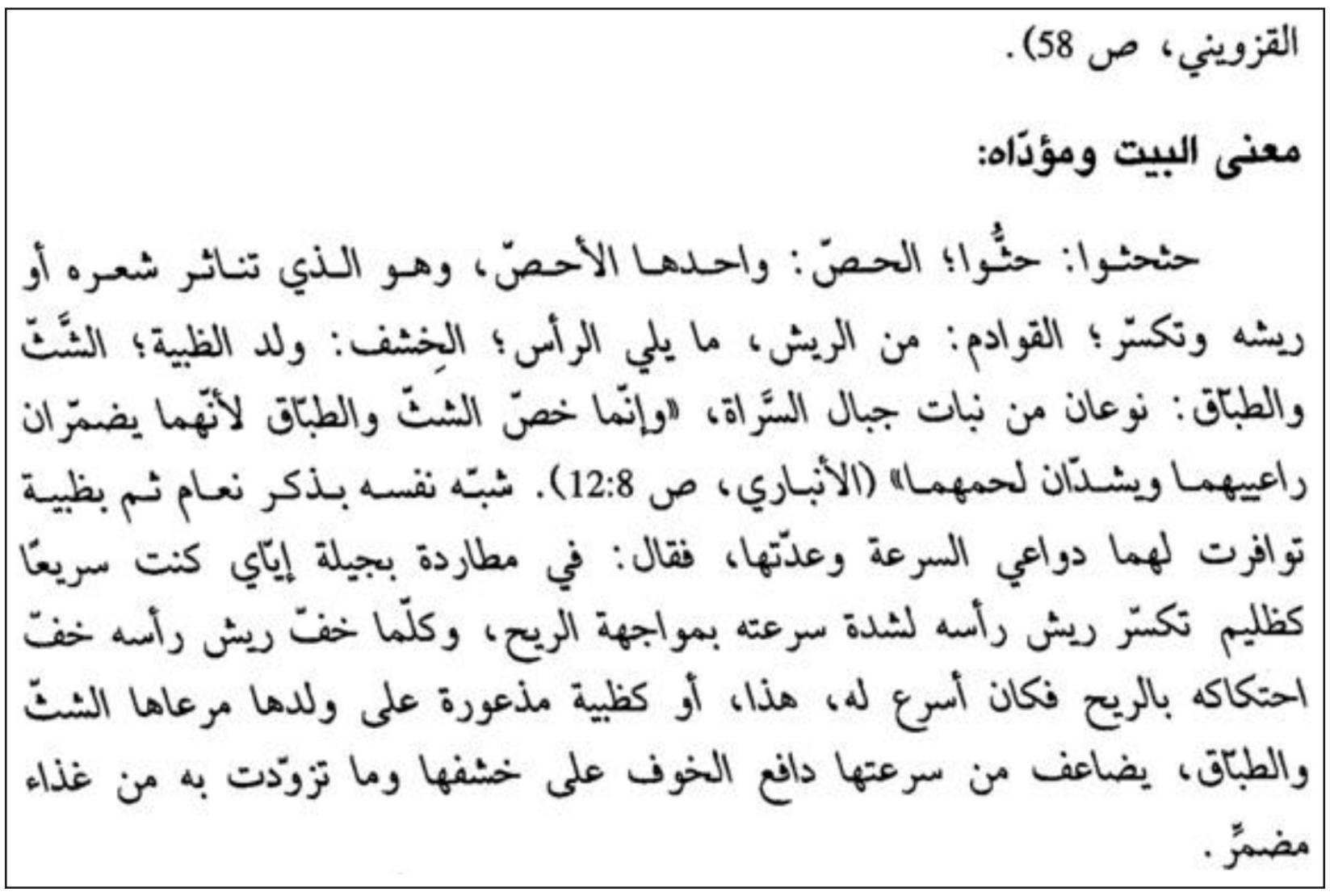
القزويني، ص 58).

معنى البيت ومؤدّاه:

حثحثوا: حثُّوا؛ الحصّ: واحدها الأحصّ، وهو الذي تناثر شعره أو ريشه وتكسّر؛ القوادم: من الريش، ما يلي الرأس؛ الخِشف: ولد الظبية؛ الشّثّ والطبّاق: نوعان من نبات جبال السَّراة، «وإنّما خصّ الشثّ والطبّاق لأنّهما يضمّران راعيهما ويشدّان لحمهما» (الأنباري، ص 12:8). شبّه نفسه بذكر نعام ثم بظبية توافرت لهما دواعي السرعة وعدّتها، فقال: في مطاردة بجيلة إيّاي كنت سريعًا كظليم تكسّر ريش رأسه لشدة سرعته بمواجهة الريح، وكلّما خفّ ريش رأسه خفّ احتكاكه بالريح فكان أسرع له، هذا، أو كظبية مذعورة على ولدها مرعاها الشثّ والطبّاق، يضاعف من سرعتها دافع الخوف على خشفها وما تزوّدت به من غذاء مضمِّر.

**Figure 5:** Example of text in italics.

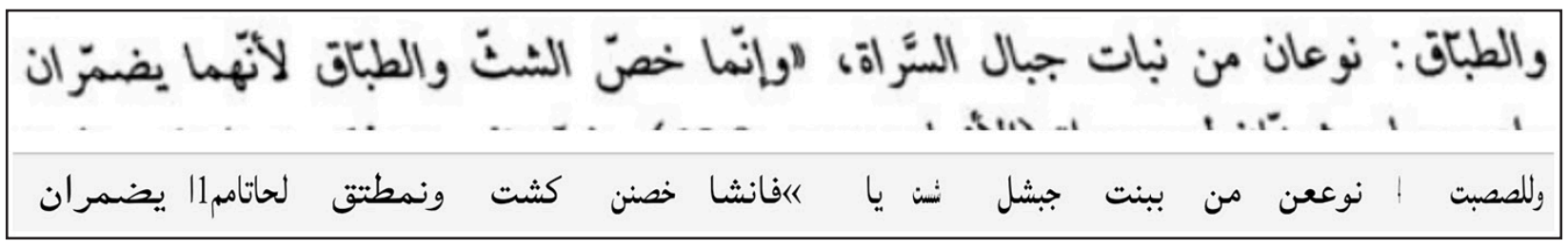
والطبّاق: نوعان من نبات جبال السَّراة، «وإنّما خصّ الشثّ والطبّاق لأنّهما يضمّران

وللصبت ا نوعمن من ببنت جبشل ئـ يا »فانشا خصنن كشت ونمطتق لحاتامII يضمران

**Figure 6:** Example of poor transcription of italicized passage in figure 5.

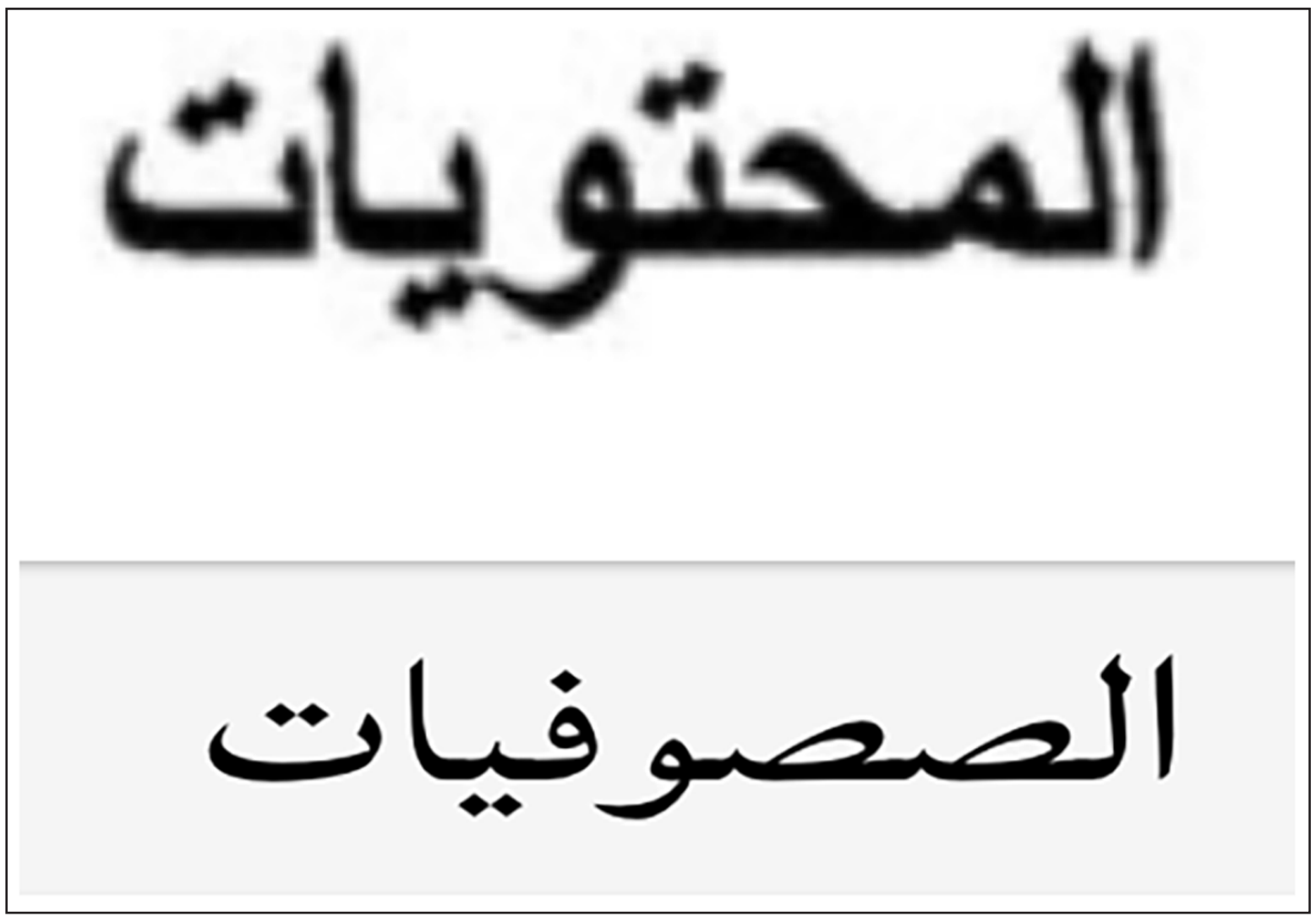

**Figure 7:** Bolded and enlarged text size header and poor transcription.

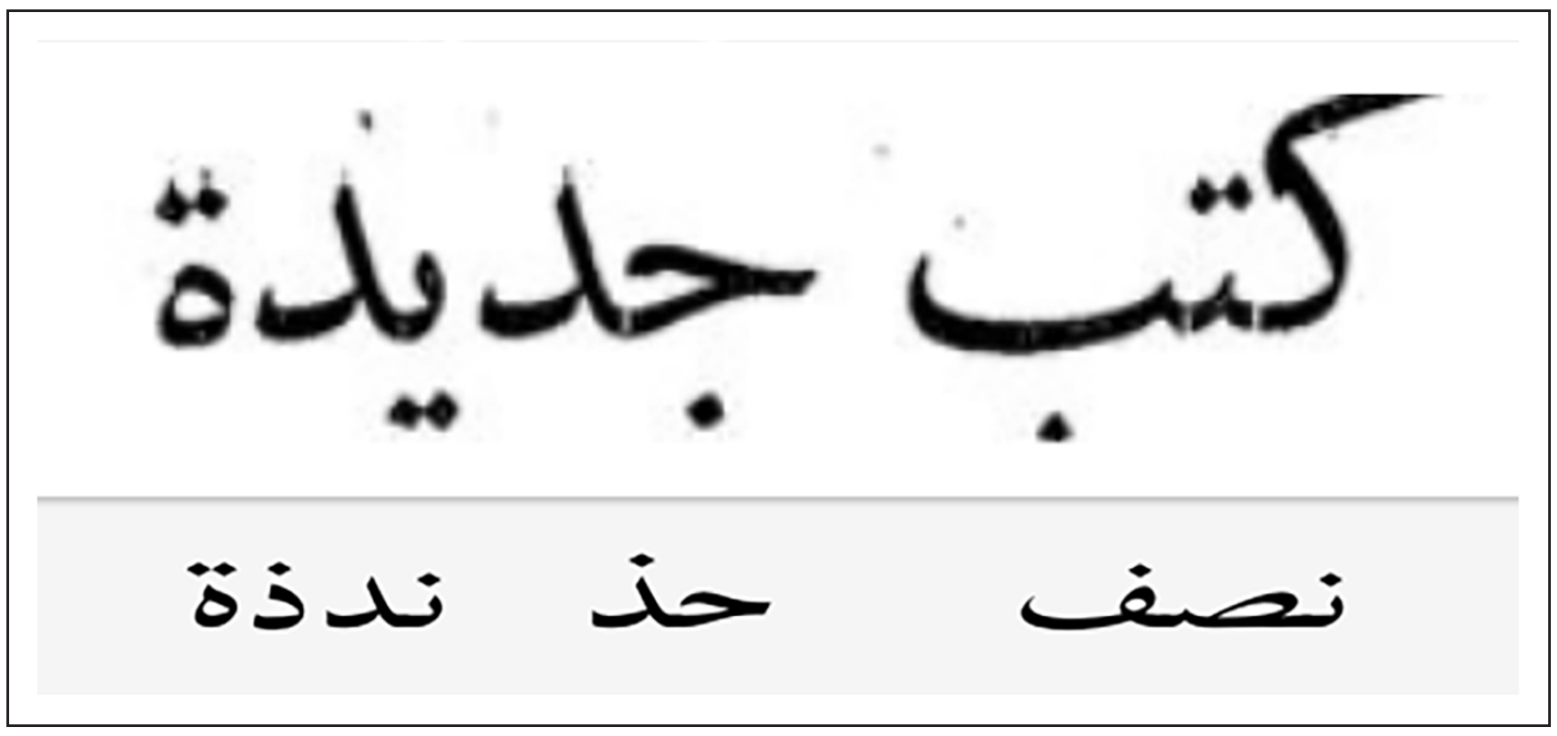

**Figure 8:** Bolded and enlarged text size header and poor transcription.

## Ligatures/Atypical Letter or dot forms

Not surprisingly, ligatures and other types of less common letter patterns and dot placements led to less accurate transcriptions in the OCR output (see **Figures 9–14**). This same problem has been observed in an earlier study as well (Kiessling et al. 2017).

It is nearly impossible to completely avoid this problem, but a more systematic approach to training data generation that selected pages/lines of data with an eye towards ensuring sufficient representation of the maximum number of ligatures could improve OCR accuracy on these characters/character combinations.

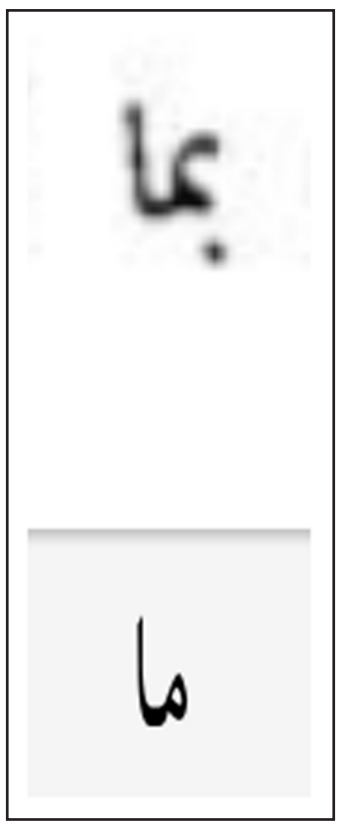

**Figure 9:** Example of problematic ligature and error in transcription.

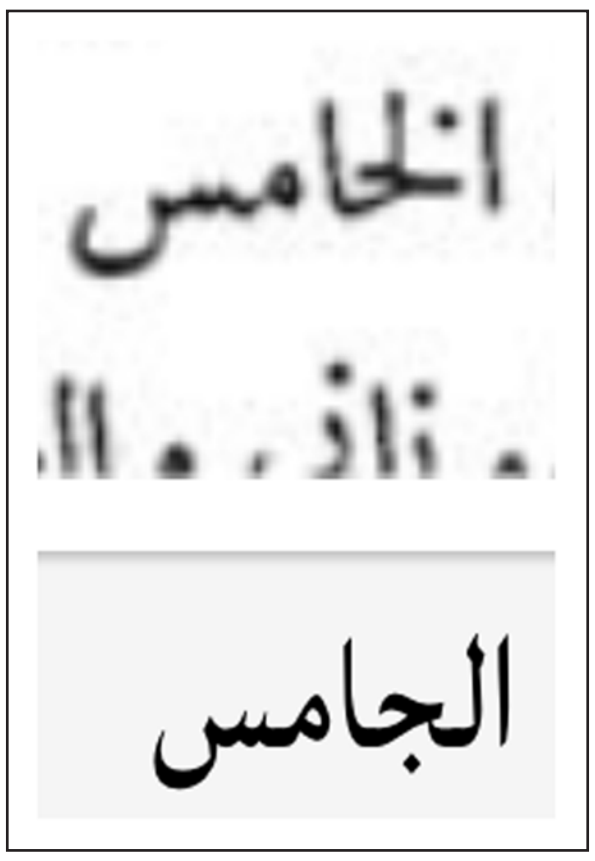

**Figure 10:** Example of problematic ligature and error in transcription.

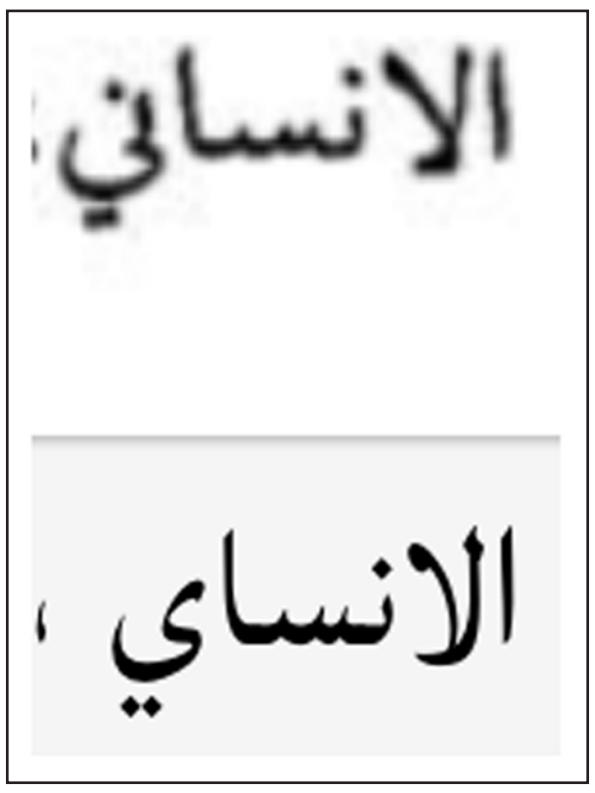

**Figure 11:** Atypical dot placement.

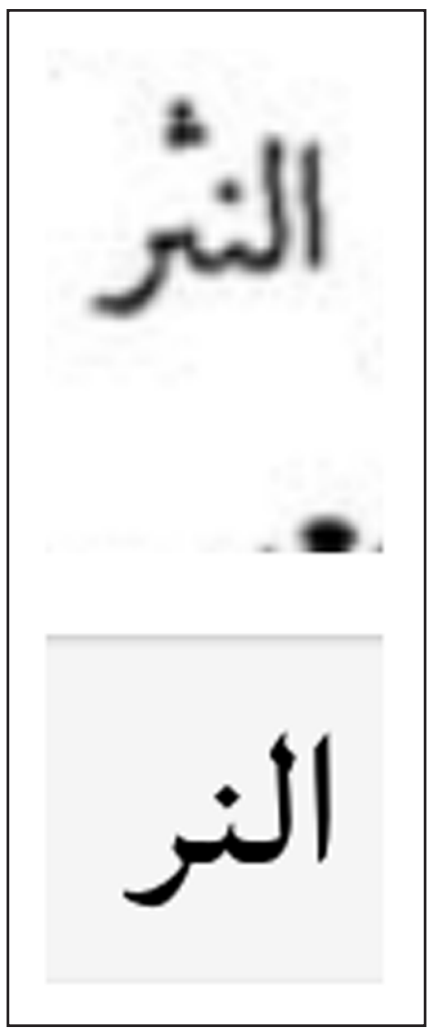

**Figure 12:** Atypical dot placement.

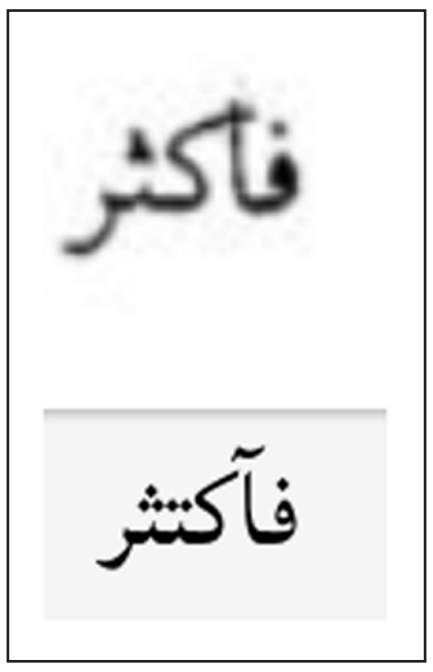

**Figure 13:** Atypical letter pattern (printing error?)

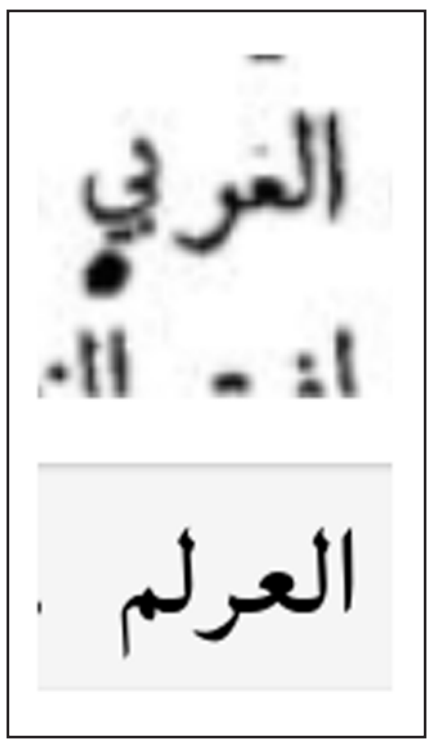

**Figure 14:** Atypical dot/letter placement and poor scan quality.

## Vocalization Diacritics

Words that contained vocalization marks also appear more frequently to have errors in transcription, which leads us to believe that they are interfering with character recognition. This tendency especially can be seen in examples of heavy vocalization, such as the fully vocalized Qur'anic passage seen in **Figure 15**, which are poorly transcribed. **Figure 15** is an extreme case that is an outlier in the *al-Abhath* data, but it clearly illustrates this problem. Moreover, although *al-Abhath* journal articles are not heavily vocalized, this could be a significant issue in other Arabic texts that are heavily vocalized.

In general, OpenITI has traditionally followed the practice of *not* transcribing Arabic vocalization marks in our training data production (with one exception discussed below). We have followed this practice for three reasons: (1) vocalization is often inconsistent and sometimes incorrect (so it is better to allow the individual scholar to determine the proper vocalization based upon their reading); (2) vocalization can interfere with computational textual analysis (computational linguists, for example, typically remove it in their normalization of texts in preparation for analysis); and, (3) not all full-text search algorithms support vocalized text in a useful way. There is one problem with this approach, however, that we have found in both this study and another concurrent one on Persian OCR. If there is a sufficient amount of diacritics in the original text, the model will "learn" to ignore vocalization marks and it will not interfere with character recognition. However, if the original text is lightly vocalized and not enough examples of vocalization marks are contained in the training data, then it appears that the model does not "learn" well enough to ignore them and thus their presence in a word interferes with accurate character recognition. This situation presents us with a dilemma around which we need to develop a set of guidelines: we do not want to include vocalization because of the aforementioned reasons and because including them in the training data will require even more time expenditure in the training data generation process, but by *not* including them in texts with light vocalization (e.g., *al-abhath*, some of the Persian texts in our other tests) character recognition is reduced in words with them.

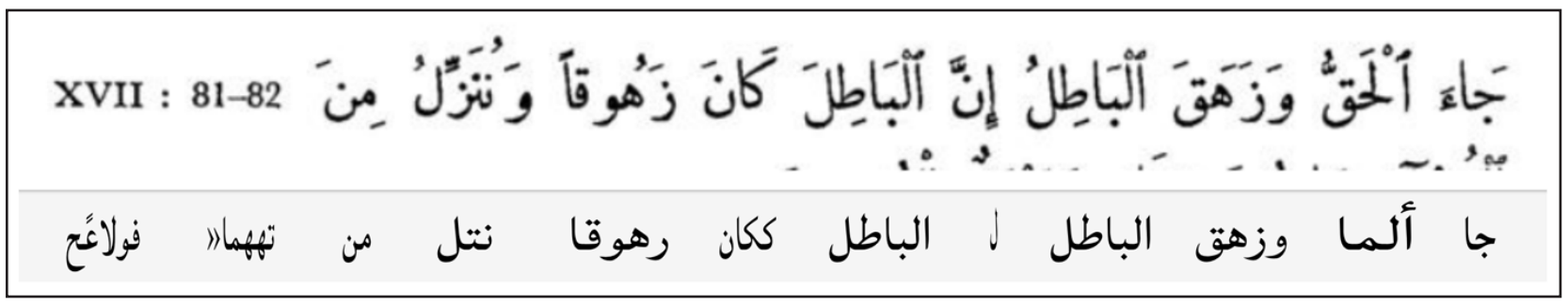

**Figure 15:** Highly vocalized Qur'anic passage that is transcribed poorly due to diacritics.

## Missed fatḥa tanwīn

The exception to our traditional treatment of vocalization discussed in the previous section is the case of the Arabic diacritic *fatḥa tanwīn* (اً). As observed in the CER reports, missed *fatḥa tanwīns* were a significant source of errors. We also observed this in the manual review of the OCR output (see **Figure 16**).

Although in the past we have not transcribed *fatḥa tanwīns* in the training data production process, we did include *fatḥa tanwīns* in the JSTOR pilot training data. In many cases the *fatḥa tanwīns* were transcribed correctly (see **Figure 17** for comparison sake). However, as both the CER reports and manual review showed, they still remained a relatively common source of errors. The reason(s) that *fatḥa tanwīn* remained a problem in the transcription process could be related to either (1) its lack of sufficient representation in the training data, or (2) its position in the line segment—i.e., in some cases it might be partially getting cut off since it appears so high in the line segment box. In either case, we are inclined to ignore *fatḥa tanwīns* in future training data production.

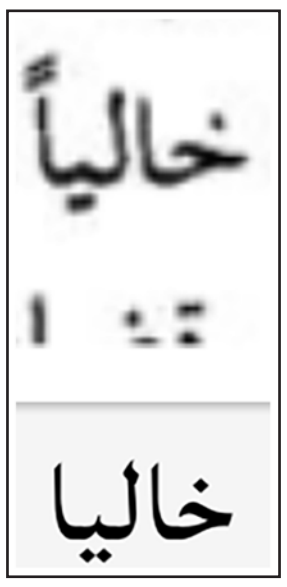

**Figure 16:** Missed fatḥa tanwīn.

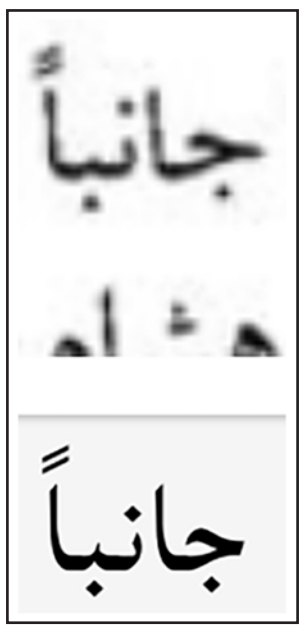

**Figure 17:** Correctly transcribed fatḥa tanwīn from same page as Figure 16.

## Punctuation marks, number, and other non-alphanumeric symbols

Punctuation marks, numbers, and other non-alphanumeric symbols (e.g., $)—especially representatives of each of these categories that were less commonly used in *al-Abhath*—were another recurring source of errors. The way to address this problem is by making sure these signs, symbols, and numbers are sufficiently represented in the training data.

## Hamzas

The *hamza* character was another common source of errors in the output, both in the sense that it was misrecognized (see **Figures 18–19**) and inserted in instances in which it was not in the original scan (see **Figure 20**).

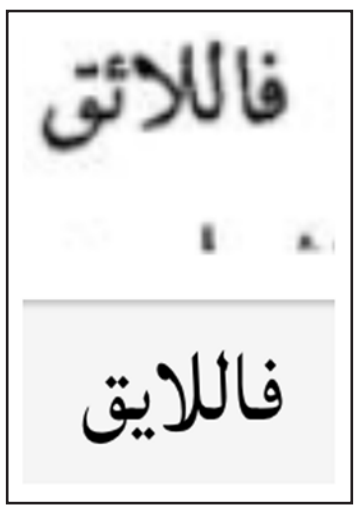

**Figure 18:** Missed hamza.

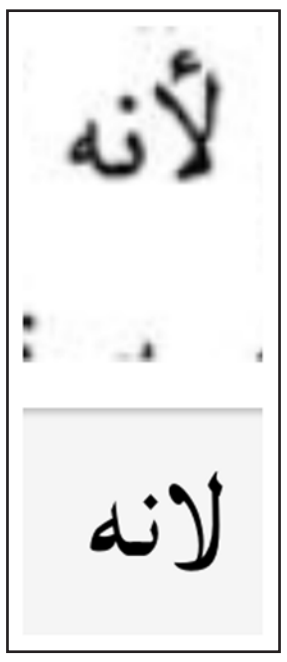

**Figure 19:** Missed hamza on alif.

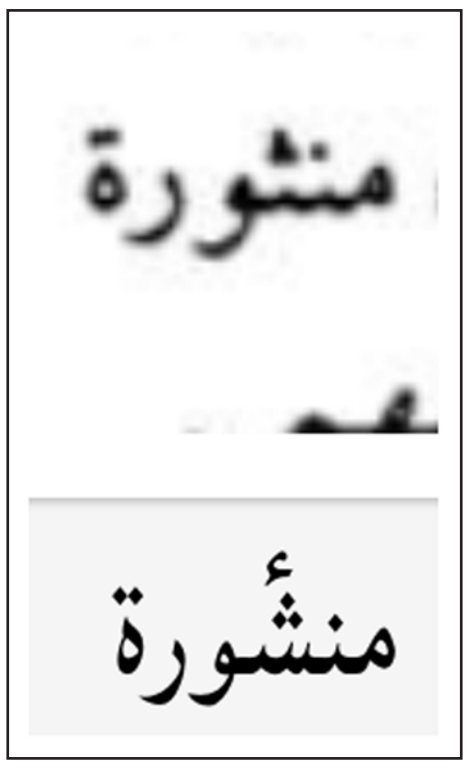

**Figure 20:** Inserted extra hamza.

Again, this is a case in which more focused training data will improve recognition rates—an intervention we must make at the training data generation phase of the OCR process.

## Atypical text presentation format and kashīda/tatwīl (elongation character)

There are a series of errors that occur in the context of atypical presentation formats/atypical character patterns. These range from the use of the Arabic elongation character (*kashīda/tatwīl*) (see **Figure 21**) to various types of table formats (see **Figures 22–24**).

Although the character recognition in these examples is usually not as poor as in **Figure 23**, we still observed that errors seem to appear more frequently in such contexts (see the better recognition in **Figures 21** and **24**). More training data from these atypical presentation formats and character patterns will help improve

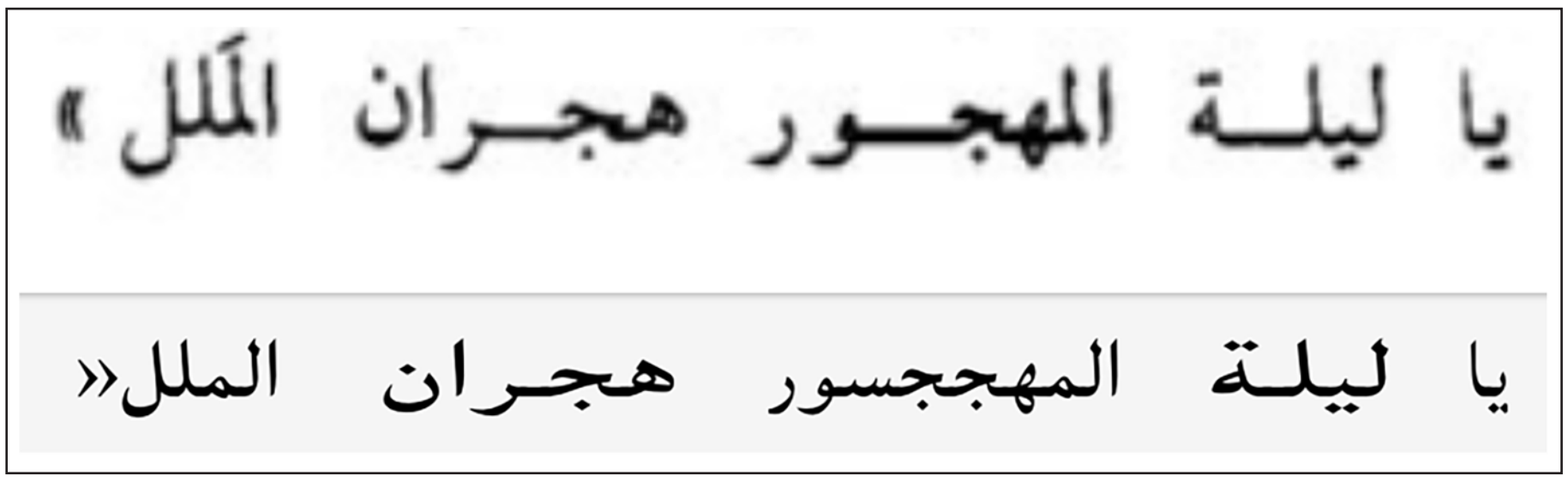

**Figure 21:** Read letter 'sin' into word due to kashīda/tatwīl (elongation).

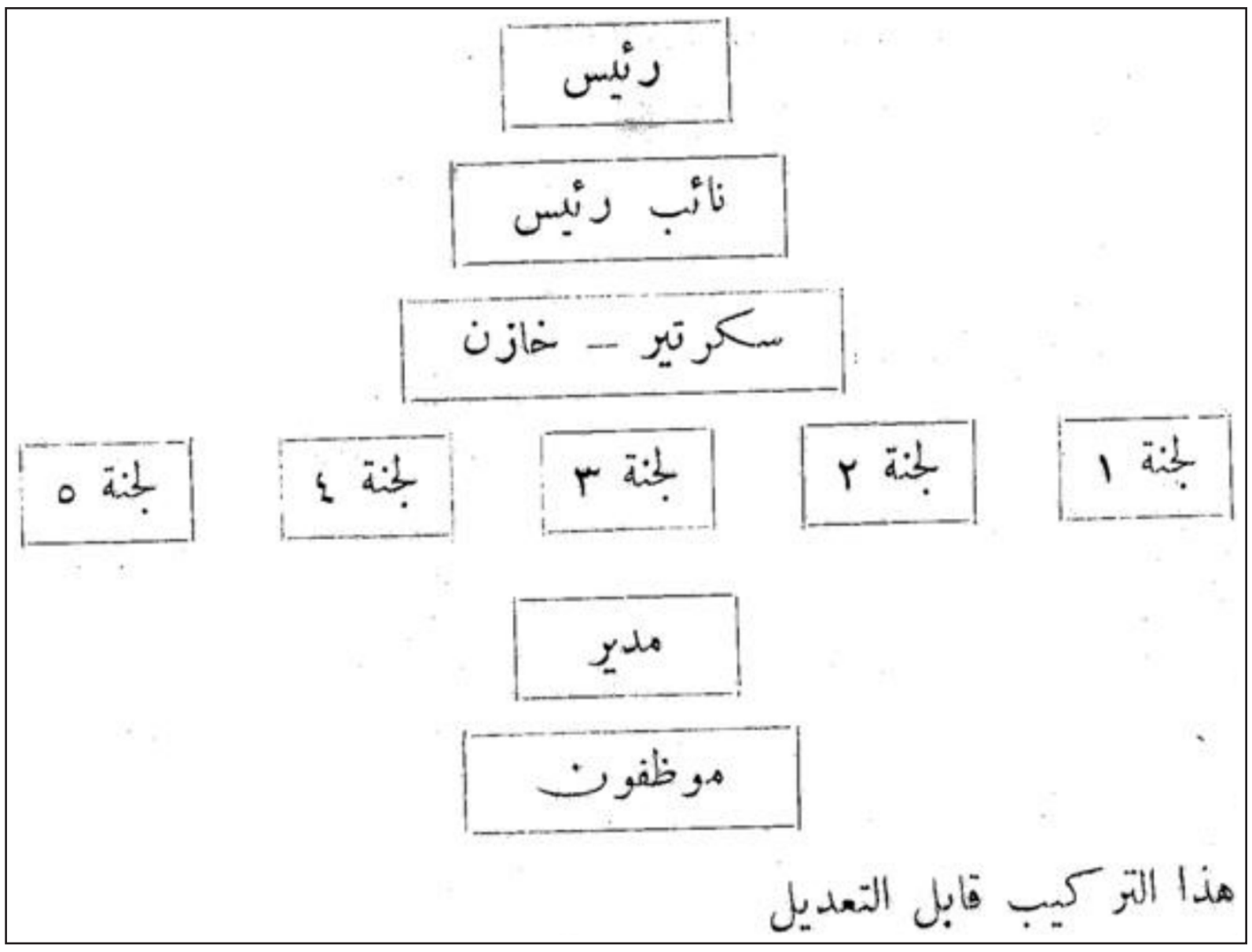

**Figure 22:** Example of table format.

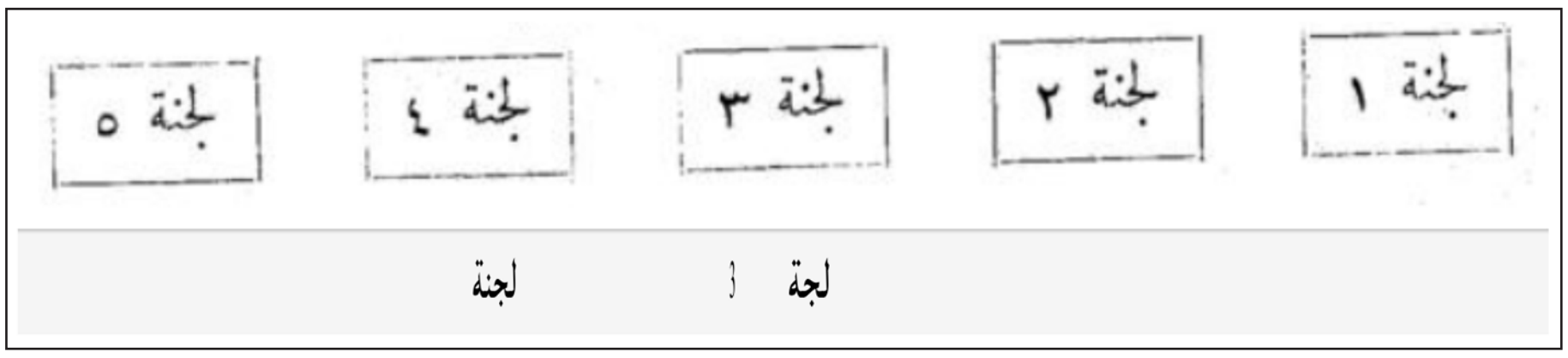

**Figure 23:** Example of particularly poor transcription on an atypical (table) presentation format.

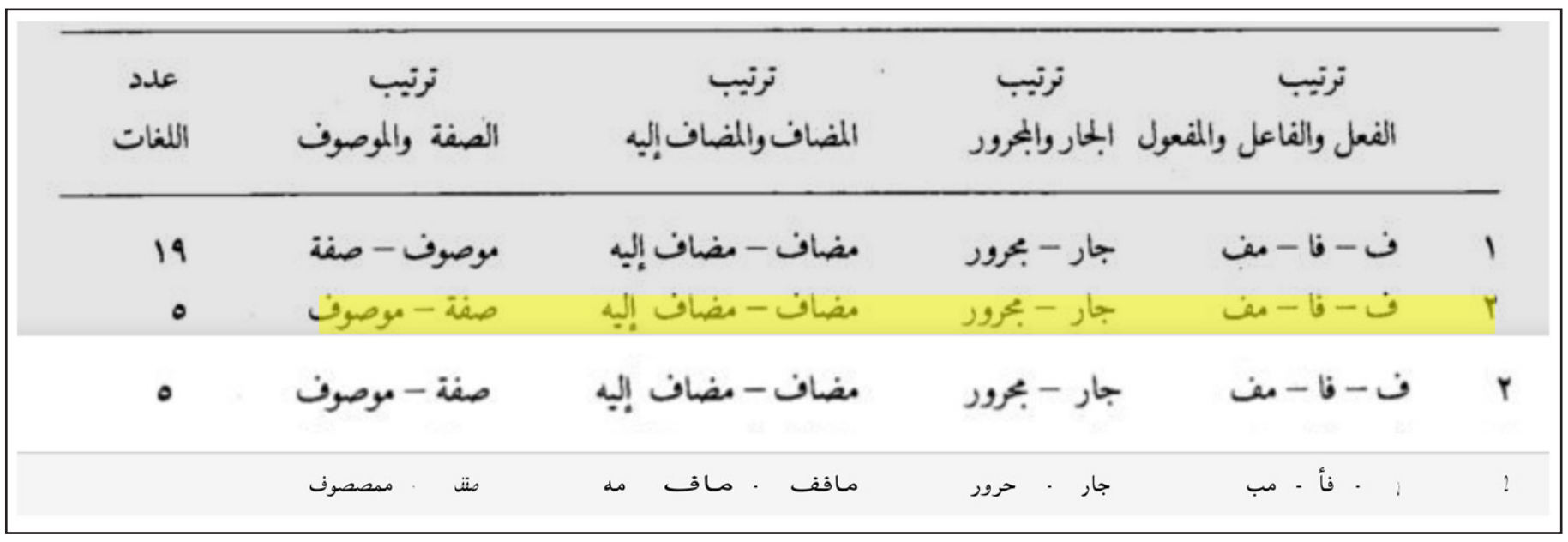

| عدد اللغات | ترتيب الصفة والموصوف | ترتيب المضاف والمضاف إليه | ترتيب الجار والمجرور | ترتيب الفعل والفاعل والمفعول | |
|---|---|---|---|---|---|
| ١٩ | موصوف – صفة | مضاف – مضاف إليه | جار – مجرور | ف – فا – مف | ١ |
| ٥ | صفة – موصوف | مضاف – مضاف إليه | جار – مجرور | ف – فا – مف | ٢ |

**Figure 24:** Example of particularly poor transcription on another atypical (table) presentation format.

accuracy, but improvements in line segmentation are also necessary for such examples as **Figure 23**.[2]

## Non-Arabic language

There were two significant types of transcription errors that were related to the presence of non-Arabic language in the original text. The first, seen in **Figure 25**, is the poor transcription of non-Arabic characters on a page that predominantly contains Arabic text. (**Figure 25** represents a particularly poor transcription of the non-Arabic text; most transcriptions in such instances were much more accurate.)

The second type of error that occurred in the context of non-Arabic script was the inverse: that is, poor transcription of Arabic text on a page that is predominantly composed of a non-Arabic language (see **Figure 26**).

This is a known problem that can be addressed through the development of multi-language OCR models—a project that OpenITI is currently working on.

## Poor scan quality

Poor scan/print quality—including, errant marks (see **Figure 27**), lack of ink (see **Figure 28**), misplaced letters/punctuation (**Figure 13**)—is not a particularly common source of errors in the *al-Abhath* data, but there is a critical mass of errors caused by this problem.

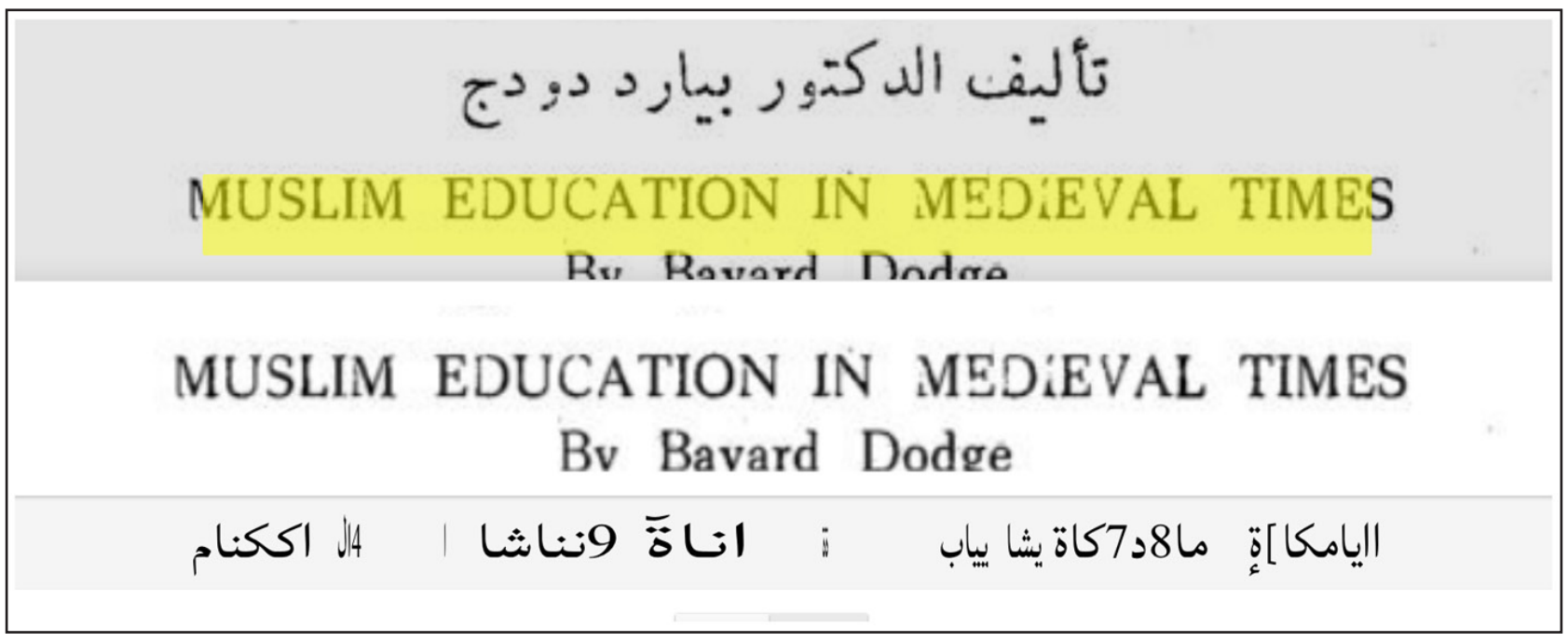

**Figure 25:** Example of particularly poor transcription of non-Arabic language in a page of primarily Arabic text.

[2] Full view of our OCR post-correction interface is shown in figures 24–26 in order to show the broader page context from which the highlighted lines are drawn and displayed in a line pair (image of line and its digital transcription) in the pop-up. Specifically, please note that in figure 24 this line is drawn from a larger table and in figures 25–26 this line is drawn from a page with significant admixture of both Arabic and Persian in the text.

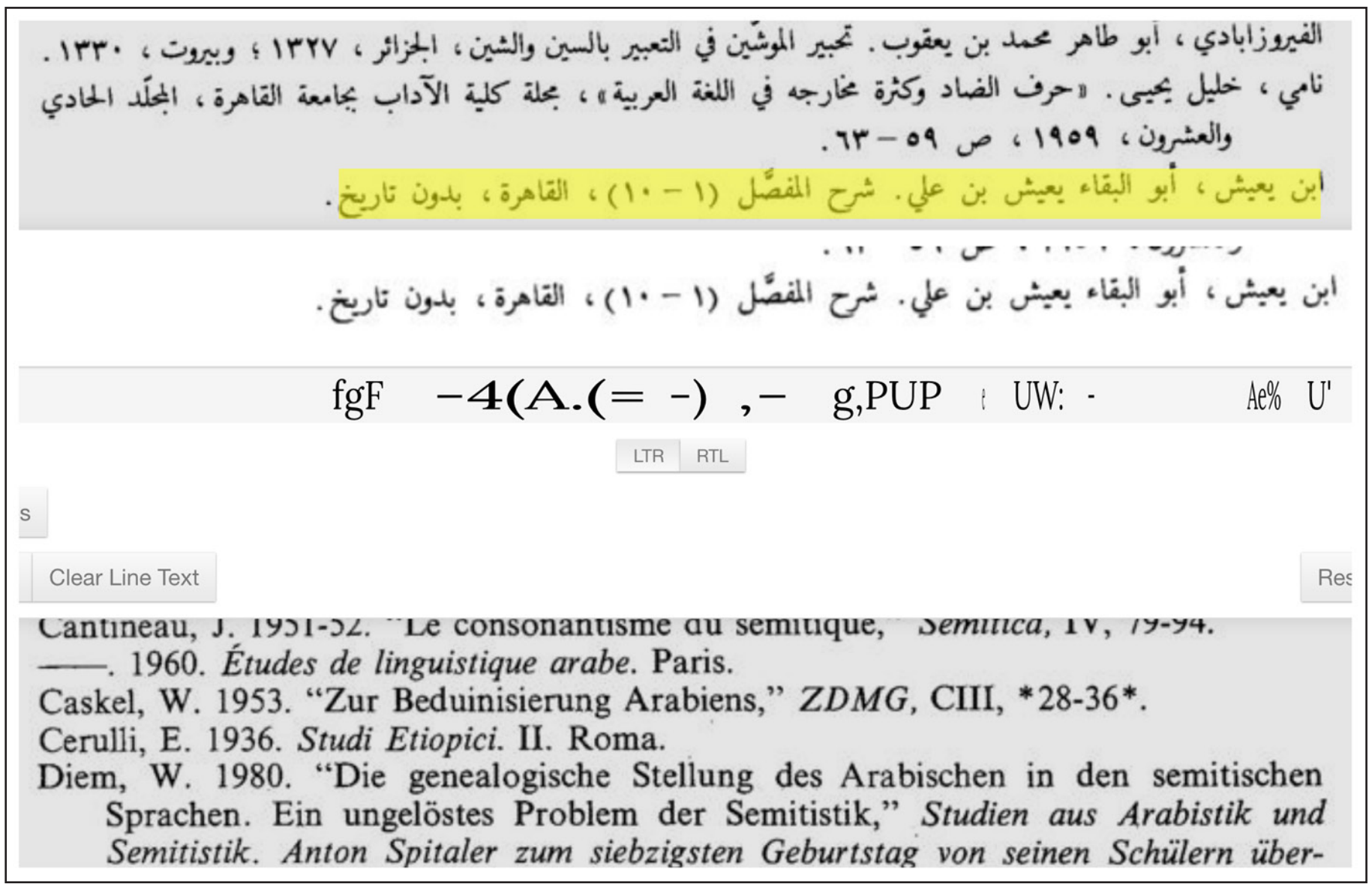

**Figure 26:** Page with substantial Non-Arabic language interferes with Arabic OCR.

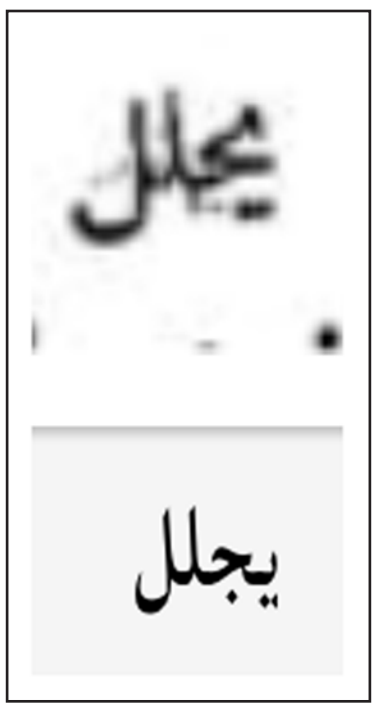

**Figure 27:** Example of poor scan quality—black shading in background of letters.

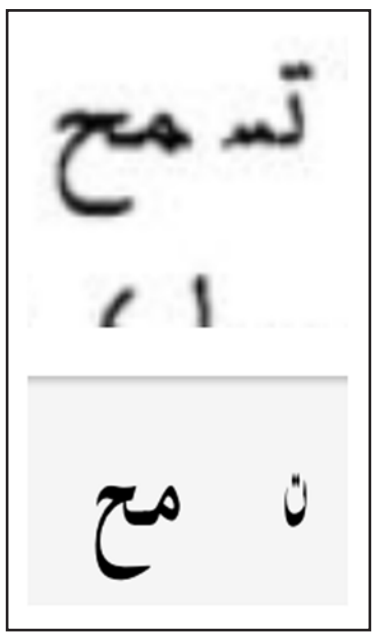

**Figure 28:** Example of poor scan quality—missing print in letter.

This problem cannot be addressed in the OCR process. OCR accuracy is (obviously) limited by the quality of the original scans.

## Line segmentation

One final error type that should be mentioned is line segmentation errors (see **Figures 29–31**).

This type of error was not commonly found in the OpenITI manual accuracy assessment (**Figures 29–30** were errors identified in the outside contractor's review of the OCR output), but there were a few cases in which the line segmenter missed a section or a word of a line. Typically this would occur in atypical text presentation formats, such as the table seen in **Figure 31**.

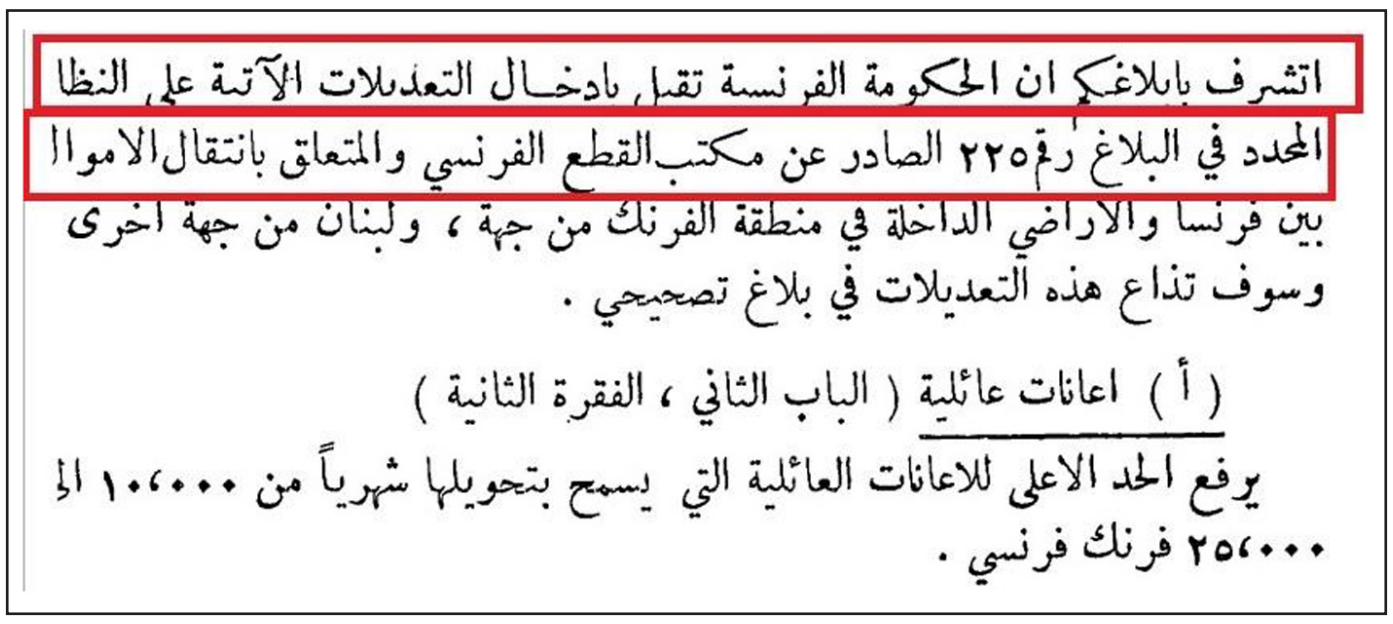
اشرف بابلاغكم ان الحكومة الفرنسية تقبل بادخال التعديلات الآتية على النظا
المحدد في البلاغ رقم ٢٢٥ الصادر عن مكتب القطع الفرنسي والمتعلق بانتقال الاموا
بين فرنسا والاراضي الداخلة في منطقة الفرنك من جهة ، ولبنان من جهة اخرى
وسوف تذاع هذه التعديلات في بلاغ تصحيحي .
( أ ) اعانات عائلية ( الباب الثاني ، الفقرة الثانية )
يرفع الحد الاعلى للاعانات العائلية التي يسمح بتحويلها شهرياً من ١٠،٠٠٠ ال
٢٥،٠٠٠ فرنك فرنسي .

**Figure 29:** Missed line segments (from outside contractor accuracy review).

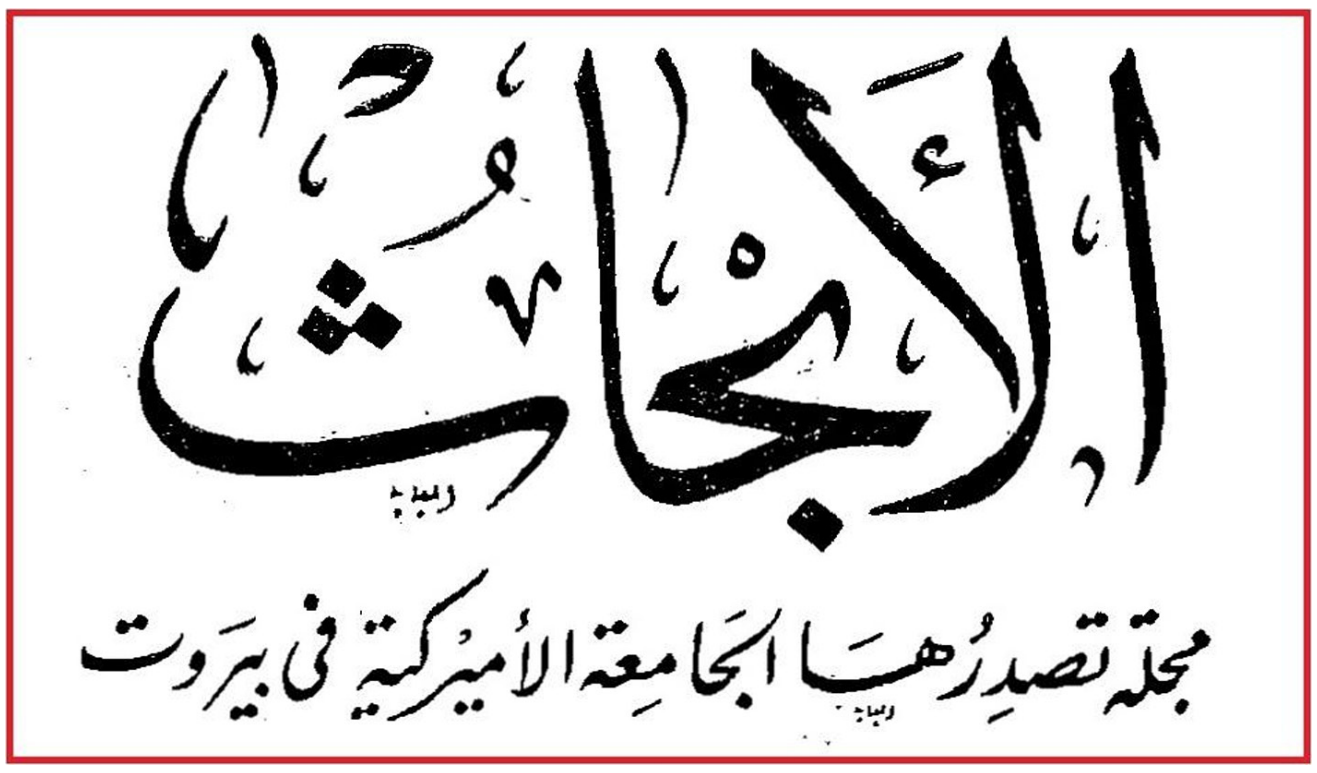

**Figure 30:** Large header segmented as one line (from outside contractor accuracy review).

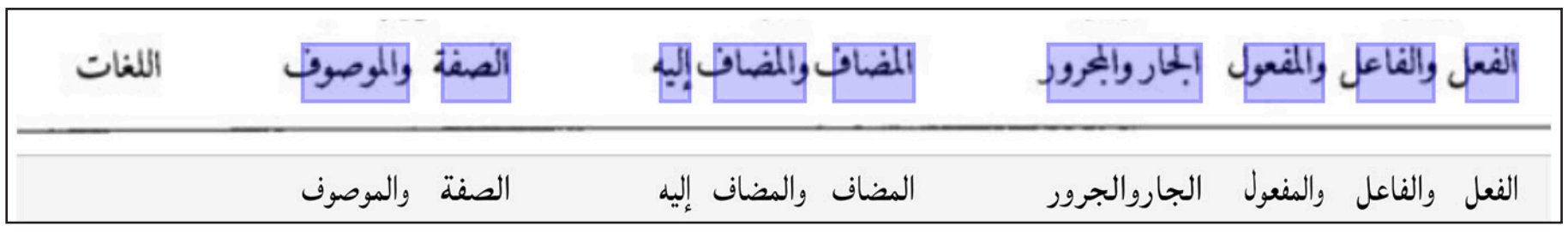

**Figure 31:** Line segmenter missed final word in the line.

Truncation of Arabic text lines is a known problem for the Latin-script-optimized line segmenter in the version of Kraken that was used for this study. The implementation of a novel trainable layout analysis method has largely solved this issue (Kiessling et al. 2019).

## Recommendations and future avenues of development for open source Arabic-script OCR

The results of this study indicate that work in the following three areas could generate significant improvements for open source Arabic-script OCR:

1. **Systematic training data production.** Instead of generating training data in a completely randomized (or haphazard) manner (as is often done), future Arabic-script OCR projects need to study the particularities of the documents they plan to OCR and make sure that the pages selected for training data production contain a sufficient number of the less common ligatures, headers, vocalization marks, footnote texts, numbers, and other particularities of the works to be OCR'd. We followed this randomized training data generation approach in the past (Kiessling et al. 2017). (See Springmann et al. 2018 for an example of a dataset resulting from haphazard convenience sampling, i.e. harvesting data from sources on which existing methods already produce near-perfect results.) This more systematic approach to training data production will require more time upfront. But the models produced in this manner could potentially achieve much higher baseline accuracy and reduce the burden of postcorrection.
2. **Generalized models.** One of the most exciting results from this study was the significant improvements in accuracy achieved with the generalized Arabic model. The success of this approach tentatively suggests that if we continue to add training data sets to this generalized model we can anticipate achieveing higher levels of accuracy on *both* typefaces on which we have already trained models *and* new typefaces for which we have no training data yet. If this pattern holds true in future studies, we would be able to gradually reduce the time and resources necessary to achieve high level accuracy (>97%) on new typefaces in the future. However, more research on generalized models is needed as both the optimal training data selection, including artificial data produced by methods such as (Milo and Gonzalez, 2019), for such models and the actual variance on an open text corpus is currently unknown.

3. There are a range of technical improvements—e.g., multi-language models, improved line segmentation and layout analysis—that could significantly improve OCR accuracy numbers. Efforts are currently underway in both the eScripta project (of which Kiessling is a team member) and OpenITI's Arabic-script OCR Catalyst Project (OpenITI AOCP) to address each of these technical issues.

## Acknowledgements
We would like to thank David Smith (Northeastern University) for his suggestions and feedback on this study. This work was supported by JSTOR through their National Endowment for the Humanities-funded feasibility study on high-quality digitization and digital preservation of Arabic scholarly journals [grant number PW-253861-17].

## Competing interests
This work was supported by JSTOR through their National Endowment for the Humanities-funded feasibility study on high-quality digitization and digital preservation of Arabic scholarly journals [grant number PW-253861-17].

## Author contributions
Authors are listed in alphabetical order. The corresponding author is mtm.

**Conceptualization**

bk, mtm

**Data curation**

bk, gk, mtm, ks

**Formal analysis**

bk

**Funding acquisition**

mtm

**Investigation**

bk, gk, mtm, ks

**Methodology**

bk, mtm

**Project administration**

mtm

**Resources**

bk, mtm

**Software**

bk

**Supervision**

mtm

**Validation**

bk

**Visualization**

bk, mtm

**Writing – original draft**

bk, mtm

**Writing – review & editing**

bk, gk, mtm, ks

## Editorial contributions

**Recommending Editor:** Dr. Daniel O'Donnell, University of Lethbridge

**Section Editor/Copy Editor:** Darcy Tamayose, University of Lethbridge Journal Incubator

**Bibliography Editor:** Shahina Parvin, University of Lethbridge Journal Incubator

## References

**Alghamdi, Mansoor,** and **William Teahan.** 2017. "Experimental Evaluation of Arabic OCR Systems." *PSU Research Review* 1(3): 229–241. United Kingdom: Emerald Publishing Limited. DOI: https://doi.org/10.1108/PRR-05-2017-0026

**Clausner, Christian, Apostolos Antonacopoulos, Nora Mcgregor,** and **Daniel Wilson-Nunn.** 2018. "ICFHR 2018 Competition on Recognition of Historical Arabic Scientific Manuscripts–RASM2018." *2018 16th International Conference on Frontiers in Handwriting Recognition (ICFHR)*, 471–476. Niagara Falls, NY. DOI: https://doi.org/10.1109/ICFHR-2018.2018.00088

**Graves, Alex, Santiago Fernández, Faustino Gomez,** and **Jürgen Schmidhuber.** 2006. "Connectionist Temporal Classification: Labelling Unsegmented Sequence Data with Recurrent Neural Networks." In *Proceedings of the 23rd International Conference on Machine Learning*, 369–376. New York: Association for Computing Machinery. Accessed April 28, 2021. DOI: https://doi.org/10.1145/1143844.1143891

**Keinan-Schoonbaert, Adi.** 2019. "Results of the RASM2019 Competition on Recognition of Historical Arabic Scientific Manuscripts." *British Library Digital Scholarship Blog*. Accessed April 28, 2021. https://blogs.bl.uk/digital-scholarship/2019/09/rasm2019-results.html.

——. 2020. "Using Transkribus for Arabic Handwritten Text Recognition." *British Library Digital Scholarship Blog*. Accessed April 28, 2021. https://blogs.bl.uk/digital-scholarship/2020/01/using-transkribus-for-arabic-handwritten-text-recognition.html.

**Kiessling, Benjamin.** 2019. "Kraken—A Universal Text Recognizer for the Humanities." *DH2019 Utrecht*. Netherlands: Utrecht University. DOI: https://doi.org/10.34894/Z9G2EX

**——.** n.d. Kraken. *GitHub Repository*. Accessed April 19, 2021. https://github.com/mittagessen/kraken.

**Kiessling, Benjamin, Daniel Stökl Ben Ezra,** and **Matthew Thomas Miller.** 2019. "BADAM: A Public Dataset for Baseline Detection in Arabic-script Manuscripts." *HIP 2019: 5th International Workshop on Historical Document Imaging and Processing*. New York: Association for Computing Machinery. DOI: https://doi.org/10.1145/3352631.3352648

**Kiessling, Benjamin, Matthew Thomas Miller, Maxim Romanov,** and **Sarah Bowen Savant.** 2017. "Important New Developments in Arabographic Optical Character Recognition (OCR)." *Al-'Usur al-Wusta* 25: 1–13. Accessed January 20, 2019. DOI: https://doi.org/10.17613/M6TZ4R

**Kiplinger, John,** and **Anne Ray.** 2019. "Digitizing Printed Arabic Journals: Is a Scalable Solution Possible?" *JSTOR Arabic-Language Digitization Planning*. Accessed April 22, 2019. https://about.jstor.org/wp-content/uploads/2019/08/NehAward_PW-253861-17_JstorArabicDigitizationInvestigation_WhitePaper_20190329.pdf

**Milo, Thomas,** and **Alicia Gonzalez Martinez.** 2019. "A New Strategy for Arabic OCR: Archigraphemes, Letter Blocks, Script Grammar, and Shape Synthesis." In *DATeCH2019 Proceedings of the 3rd International Conference on Digital Access to Textual Cultural Heritage*, 93–96. Brussels, Belgium.

**Open Islamicate Texts Initiative.** 2021a. *CorpusBuilder 1.0*. Accessed April 19. https://www.openiti.org/projects/corpusbuilder.

———. 2021b. "OCR_GS_Data." *GitHub Repository*. Accessed April 19. https://github.com/OpenITI/OCR_GS_Data/tree/master/fas.

**Springmann, Uwe, Christian Reul, Stefanie Dipper,** and **Johannes Baiter.** 2018. "Ground Truth for Training OCR Engines on Historical Documents in German Fraktur and Early Modern Latin." In *Journal for Language Technology and Computational Linguistics* 33(1): 97–114. Germany: German Society for Computational Linguistics and Language Technology.